\documentclass[sigconf]{acmart}

\renewcommand\footnotetextcopyrightpermission[1]{}
\usepackage{booktabs}
\usepackage{balance}
\usepackage{amsmath}
\usepackage{verbatim}
\usepackage{amssymb}
\usepackage{algorithm,algpseudocode}
\usepackage{color}

\def\ie{\textit{i.e.}}
\def\eg{\textit{e.g.}}

\def\etal{\textit{et~al.~}}

\begin{document}
\title{Precise Temporal Action Localization by\\ Evolving Temporal Proposals}

\author{Haonan Qiu}
\authornote{These authors contributed equally to this work.}
\affiliation{%
  \institution{East China Normal University}
  \city{Shanghai}
  \country{China}
}
\email{hnqiu@ica.stc.sh.cn}

\author{Yingbin Zheng}
\authornotemark[1]
\affiliation{
  \institution{Shanghai Advanced Research Institute, CAS, China}
}
\email{zhengyb@sari.ac.cn}

\author{Hao Ye}
\authornote{Corresponding author.}
\affiliation{
  \institution{Shanghai Advanced Research Institute, CAS, China}
}
\email{yeh@sari.ac.cn}

\author{Yao Lu}
\affiliation{
  \institution{University of Washington}
  \city{Seattle, WA}
  \country{USA}
}
\email{luyao@cs.washington.edu}

\author{Feng Wang}
\affiliation{
  \institution{East China Normal University}
  \city{Shanghai}
  \country{China}
}
\email{fwang@cs.ecnu.edu.cn}

\author{Liang He}
\affiliation{
  \institution{East China Normal University}
  \city{Shanghai}
  \country{China}
}
\email{lhe@cs.ecnu.edu.cn}

\begin{abstract}
Locating actions in long untrimmed videos has been a challenging problem in video content analysis. The performances of existing action localization approaches remain unsatisfactory in precisely determining the beginning and the end of an action. Imitating the human perception procedure with observations and refinements, we propose a novel three-phase action localization framework. Our framework is embedded with an Actionness Network to generate initial proposals through frame-wise similarity grouping, and then a Refinement Network to conduct boundary adjustment on these proposals. Finally, the refined proposals are sent to a Localization Network for further fine-grained location regression. The whole process can be deemed as multi-stage refinement using a novel non-local pyramid feature under various temporal granularities. We evaluate our framework on THUMOS14 benchmark and obtain a significant improvement over the state-of-the-arts approaches. Specifically, the performance gain is remarkable under precise localization with high IoU thresholds. Our proposed framework achieves mAP@IoU=0.5 of 34.2\%.

\end{abstract}

\keywords{Action localization; temporal proposal; deep neural network}

\maketitle

\begin{figure}[t]
  \centering
  \includegraphics[width=0.85\linewidth]{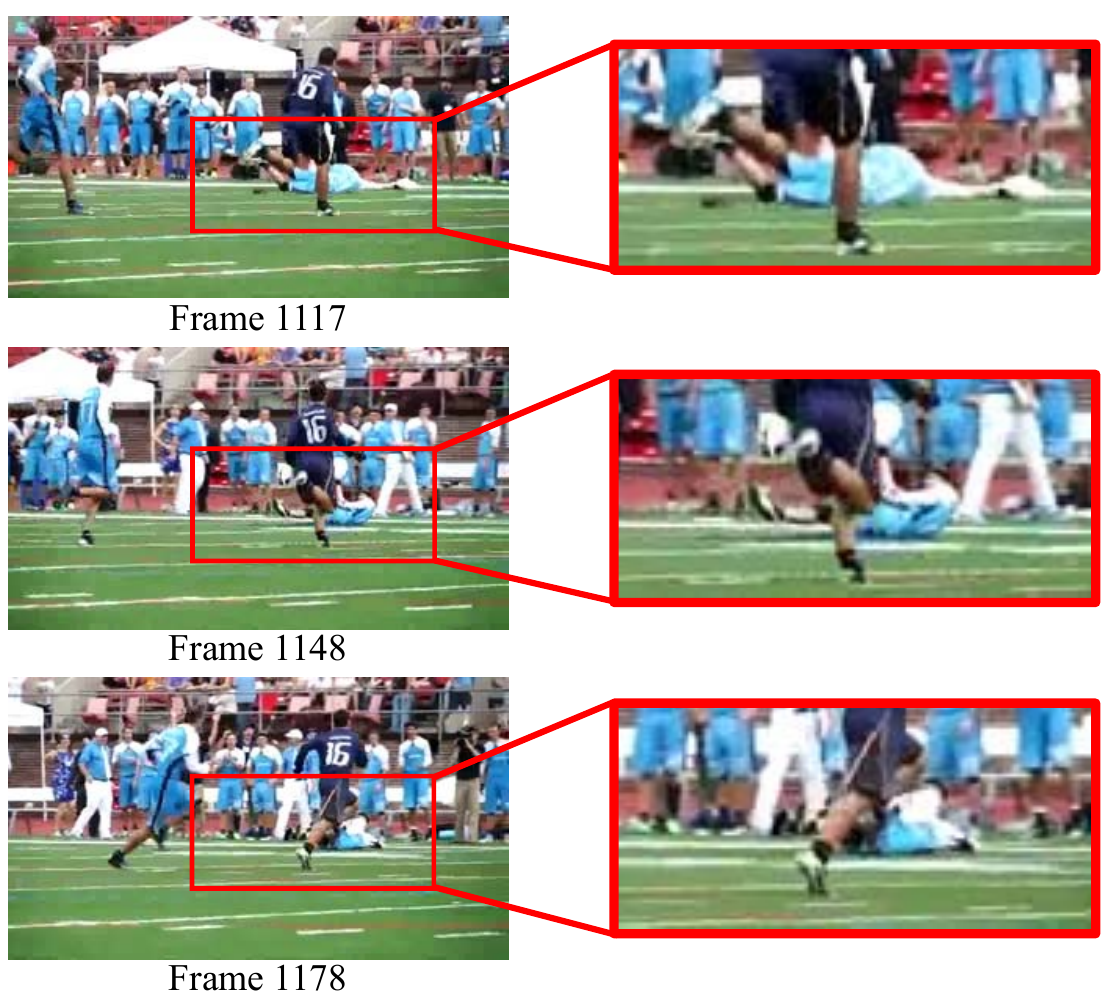}
  \caption{Localization of precise action boundary is challenging. We demonstrate an example from the THUMOS14 dataset \cite{jiang2014thumos}, and the middle frame is the annotated ending of the action ``FrisbeeCatch''.}
  \label{fig:boundary}
\end{figure}

\section{Introduction}
\label{sec:intro}

With the ubiquitous video capture devices, a huge amount of videos are producted, which brings a great demand for automatic video content analysis \cite{lu2016optasia,2010mm_yzheng,simonyan2014two,ye2015evaluating,2017iccv_tap,2017arxiv_kyang,shou2016temporal,yeung2016end,shou2017cdc,yuan2016temporal,Zhao_2017_ICCV,jiang2013understanding}. Action localization and recognition answer \textit{where} and \textit{what} an action is in the video. Since most collected videos are untrimmed, action localization becomes, undoubtedly, the first and the foremost steps in video action analysis.
On common image object detection, the ``\textit{Proposal} + \textit{Classification}'' framework demonstrates its capability  \cite{girshick2014rich,ren2015faster,2016eccv_wliu,2016cvpr_JRedmon}, and is easy to be transferred to other detection tasks, such as detecting the face \cite{2017fg_hjiang}, text \cite{ma2017arbitrary} and vehicle \cite{wang2017eb}.
For the video action localization task, the high-quality temporal proposals are also crucial following this paradigm.
A promising temporal proposal candidate in action localization should contain the action of interest in accordance with high Intersection-over-Union (IoU) overlap with the groundtruth.

Recently, training deep neural networks to extract the spatial-temporal features of the proposals is widely used in action localization. Both the regressor and the classifier are trained to determine the fidelity of the boundary and the completeness of an action based on these proposal features. Nevertheless, regressing the boundary and judging the completeness of an action are nontrivial tasks due to two difficulties. First, compared to the object proposals which have solid internal consistency, the boundary between an action and the background is vague, because the variations of the consecutive video frames are subtle. This may lead to an unsteady or even incorrect boundary regression, especially under frame-level granularity. Second, it is very subjective to judge the completeness of an action from the background. The action is usually complex and diverse, which makes it hard to discern action snippets from backgrounds. Figure~\ref{fig:boundary} illustrates an example of the unclear boundary between the action and the background in some action videos. These two issues impede the accuracy of action localization in the long untrimmed videos.

To precisely detect the actions in the untrimmed videos, many efforts have been made to generate well-anchored temporal proposals and to judge the actionness of the proposals. Grouping sampled snippets of various granularities in a bottom-up manner can generate proposals with vastly varying lengths (\eg, in  \cite{yuan2016temporal,2017arxiv_yxiong}). However, snippets grouping is much dependent on the selections of the similarity metrics and the threshold, which decides the clearness of the action boundaries.
In this paper, we propose an evolving framework in which the boundary of temporal proposals is precisely predicted with a three-phase algorithm. The first phase generates temporal proposals by an Actionness Network (AN) which employs frame-level features. The goal is to discern the actionness of the proposals. The second phase refines the boundaries of these proposals through a Refinement Network (RN) to improve the fidelity of the proposal boundaries. It turns out that coarse-grained temporal boundary regression is more effective and stable. That is, by segmenting a proposal into smaller units, RN adjusts the boundary on these unit-based features. Finally, the adjusted proposals are sent to the third phase, namely Localization Network (LN), to precisely refine the action boundary. With our evolving framework, we demonstrate a significant improvement over those regressed directly from the original proposals.

Our contributions are summarized as follows.
\begin{itemize}
 	\item We propose a novel three-phase evolving temporal proposal framework for action localization using multi-stage temporal coordinate regression under various temporal granularities.
 	\item We exploit unit-based temporal coordinate regression to the boundaries of the proposals to precisely locate the action.
 	\item We use non-local pyramid features to effectively model an action, which is capable of discriminating between completeness and incompleteness of a proposal.
 	\item Our proposed framework outperforms the state-of-the-arts approaches, especially in precise action localization.
\end{itemize}

The remaining of this paper is organized as follows. Section \ref{sec:related} briefly reviews the literature on temporal action localization and the related topics. Section \ref{sec:app} presents the details of the evolving architecture of our framework. We empirically evaluate our proposed framework in Section \ref{sec:exp}. Finally, Section \ref{sec:conclusion} concludes this paper.

\section{Related Works}
\label{sec:related}

\subsection{Action Recognition}
Action recognition has been widely studied during last few years.
Early works focus on hand-crafted feature engineering. Various features have been invented, for instance, space-time interest points (STIP)~\cite{laptev2005space} and histogram of optical flow (HOF)~\cite{laptev2008learning}. With the development of deep neural network, more effective features are extracted using various deep neural networks. Multi-stream features are fused for action recognition and show remarkable performance~\cite{simonyan2014two,ye2015evaluating,wu2016multi}.

\subsection{Temporal Action Proposal}
Temporal action proposal is the elementary factor in the ``\textit{Proposal} + \textit{Classification}''
paradigm.
Escorcia~\etal use LSTM networks to encode a video stream and produce the proposals inside the video stream~\cite{escorcia2016daps}.
Buch \etal develop a new effective deep architecture to generate single-stream temporal proposals without dividing the video into short snippets~\cite{Buch:2017dk}.
Gao~\etal use unit-level temporal coordinate regression to predict the temporal proposals and refine the proposal boudnaries~\cite{2017iccv_tap}.
Gao~\etal exploit cascaded boundary regression to refine the sliding windows as class-agnostic proposals~\cite{2017arxiv_jgao}.
Xiong~\etal propose a learning-based bottom-up proposal generation scheme called temporal actionness grouping (TAG)~\cite{2017arxiv_yxiong}.

\begin{figure*}[t]
  \centering
  \includegraphics[width=.9\linewidth]{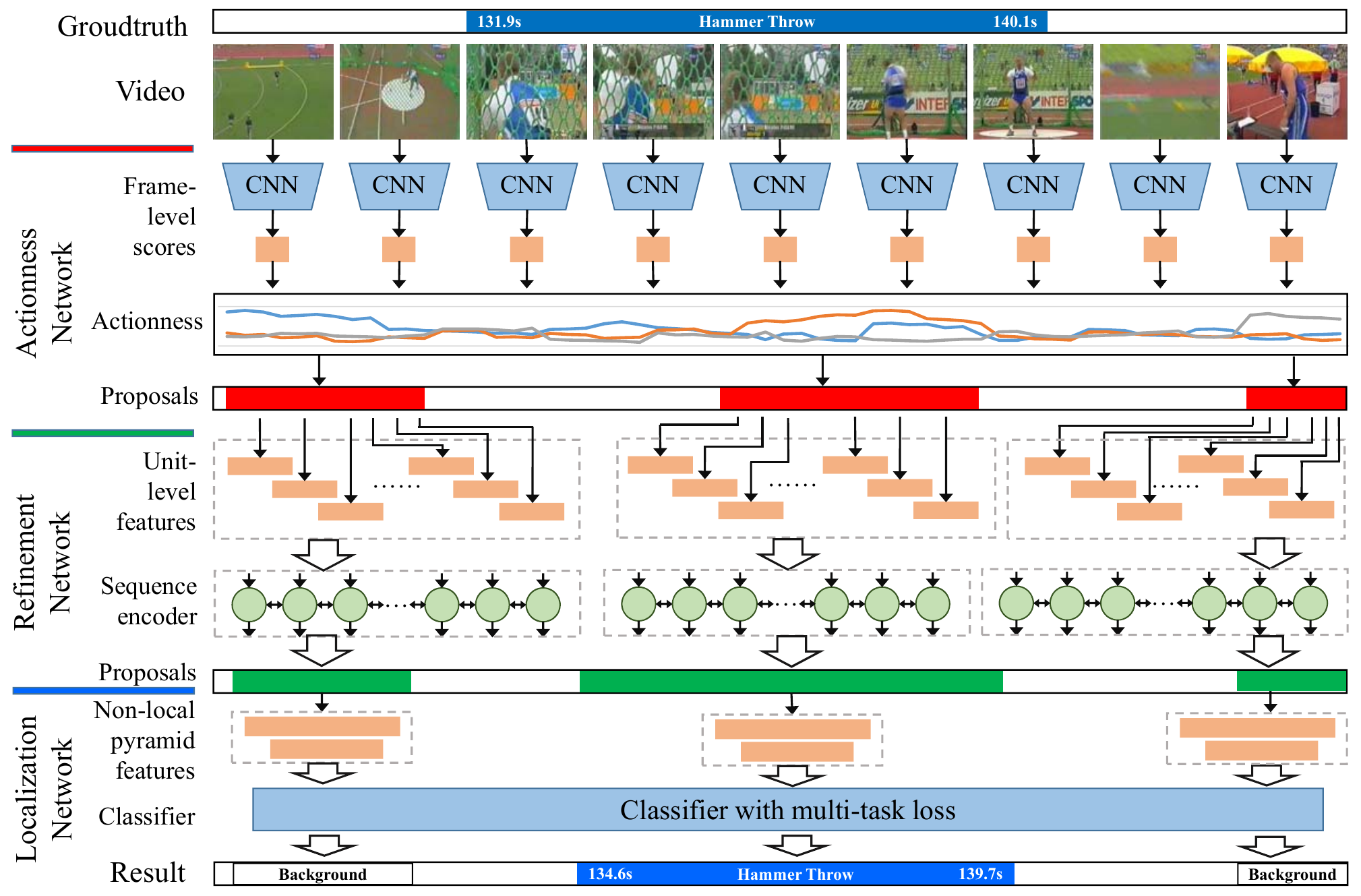}
  \caption{The evolving architecture of our framework. Given a video, we use the Actionness Network to generate initial temporal action proposals. For each proposal, a few video units are extracted and represented by the unit-level features. The features are input to the GRU-based sequence encoder, which produces a refined temporal proposal. The predictions of all proposals are computed in the Localization Network with the non-local pyramid features.}
  \label{fig:framework}
\end{figure*}

\subsection{Temporal Action Localization}
Action localization aims to predict where an action begins and ends in the untrimmed videos.
Shou~\etal propose a Segment-CNN (S-CNN) proposal network and address temporal action localization by using 3D ConvNets (C3D) features, which involve two stages, namely proposal network and localization network~\cite{shou2016temporal}.
Xu~\etal introduce the regional C3D model which uses a three-dimensional fully convolutional network to generate the temporal regions for activity detection~\cite{Xu_2017_ICCV}.
Yeung~\etal propose an end-to-end learning framework using LSTM and CNN features separately, and reinforcement learning for action localization. However, each module is sub-optimal and the action boundaries are not predicted accurately~\cite{yeung2016end}.
Shao~\etal extend two-stream framework to multi-stream network and uses bi-directional LSTM network to encode temporal information for fine-grained action detection~\cite{singh2016multi}.
Ma~\etal address the problem of early action detection. That is, to train an LSTM network with ranking loss and merge the detection spans based on the frame-wise prediction scores generated by LSTM~\cite{ma2016learning}.
Yang~\etal propose temporal preservation convolutional network (TPC) which preserves temporal resolution but down-samples the spatial resolution simultaneously to detect the activities~\cite{2017arxiv_kyang}.
Zhao~\etal propose structured segment network to model activities via structured temporal pyramid~\cite{Zhao_2017_ICCV}.
Yuan~\etal introduce a pyramid of score distribution feature to capture the motion information for action detection~\cite{yuan2016temporal}.

\section{Evolving Temporal Proposals}
\label{sec:app}

We illustrate the evolving architecture and networks of our deep learning framework in Figure \ref{fig:framework}, which are composed of the following components.
\begin{itemize}
\item \emph{Actionness Network} (AN). Each frame of the videos is firstly fed to the network to compute frame-level scores. These actionness scores are used to generate initial proposals.
\item \emph{Refinement Network} (RN). This network takes the proposals from AN as input. For a single proposal, several video units are cropped with a fixed stride and length. The unit-level features are extracted and fed into a GRU-based sequence encoder. After connecting to a fully connected layer with the target of proposal position, the corresponding refined temporal proposals are produced.
\item \emph{Localization Network} (LN). The LN takes the refined proposals from the previous network as input. Non-local pyramid features are extracted from the proposals and are sent to the classifier with a multi-task loss. The final outputs are the position of evolved temporal proposals and the appropriate scores.
\end{itemize}
We describe below each individual component in detail.

\subsection{Actionness Network}

\begin{algorithm}[t]
    \caption{Proposal generation in AN}
    \begin{algorithmic}[1]
        \State {\textbf{Input}}: Action predict score vectors $S^1,S^2,...,S^K$, minimum (maximum) length $min$ ($max$), threshold $t$
        \State {\textbf{Output}}: Candidate Proposals Set $CPSet$
        \Function {Proposal}{$(S^1,S^2,...,S^K),min,max,t$}
            \State Average score vector $S^{K+1}=\frac{1}{K}\sum_{i=1}^KS^i$
            \For {$k \in$ 1:$K+1$}
            \State $PSet_k\leftarrow $\textproc{ConnComponent}($S^k,min,max,t$)
            \State $\hat{S}^k\leftarrow $Smooth($S^k$)
            \State $PSet'_k\leftarrow $\textproc{ConnComponent}($\hat{S}^k,min,max,t$)
            \State Merge $PSet_k$,$PSet'_k$ into $CPSet$
            \EndFor
            \State $CPSet\leftarrow $ NMS($CPSet$)
            \State \Return $CPSet$
        \EndFunction
        \Function {ConnComponent}{$S,min,max,t$}
            \State Candidate Set $CSet\leftarrow\{\{i\}|S_i\geq t\}$
            \State Proposal Set $PSet\leftarrow\emptyset$
            \While {$|CSet|>1$}
            \For {$C_i \in CSet~(C_i=\{p,...,q\})$}
            \State $C_i=\{p-1,p,...,q,q+1\}$
            \EndFor
            \For {each pair $(C_i,C_j) \in CSet$}
            \If {$C_i\cap C_j\neq \emptyset$}
            \State Remove $C_i,C_j$ from $CSet$
            \State Expand $C_i\cup C_j$ to $CSet$
            \EndIf
            \EndFor
            \For {$C \in CSet$}
            \State Move $C$ to $PSet$, {\bf if} $min<|C|<max$
            \EndFor
            \EndWhile
            \State \Return $PSet$
        \EndFunction
    \end{algorithmic}
    \label{alg:iou}
\end{algorithm}

Initial temporal action proposals are produced by this network. \emph{Actionness} is introduced and discussed in \cite{2017arxiv_yxiong} as the measure of possibility of a video snippet residing in any activity instance. The concept of actionness is similar to the saliency measurement that indicates the salient visual stimuli for a given image%~\cite{Itti2001Computational,2006nips_jharel,2009ICCV_TJudd,2010CVPR_balexe,Cheng2014BING}.
~\cite{Itti2001Computational,2006nips_jharel,2010CVPR_balexe,lu2011salient,Cheng2014BING,lu2012learning,lu2017closing}.
Features used for learning a model of saliency include the low-level features (\eg, orientation and color), mid-level features (\eg, surface and horizon line), and high-level features (\eg, face and objects). The saliency can be divided into \emph{generic} saliency (the degree of uniqueness of the neighborhood w.r.t. entire image) or \emph{class-specific} saliency (the visual characteristics that best distinguish a particular object class from others). For the actionness, previous work \cite{2017arxiv_yxiong} handles the generic case by learning the binary classifier. Due to the limited size of the video dataset versus the complexity of the concept actionness, only using a binary classifier may not achieve ideal classification results. Therefore, in this paper we focus on the frame-level class-specific actionness. A classification network is trained based on the frames from the UCF101 dataset~\cite{soomro2012ucf101}. This step accepts arbitrary networks designed for image classification task, and actionness results may benefit from an effective structure. Here we leverage the ResNet architecture~\cite{2016cvpr_khe} by first loading the weights pre-trained on the ImageNet dataset~\cite{ILSVRC15} and then fine-tuning on the UCF101 dataset~\cite{soomro2012ucf101}. We denote a video as $X_v = \{x_t|t=1,...,T\}$, where $T$ is the number of frames and $x_t$ is the $t$-th frame. $x_t$ id fed to the classifier to get the action prediction score. The predicted scores from all the frames are stacked and form the action score vectors $S^1,S^2,...,S^K$, where $S^k=[s_1^k,...,s_T^k]$ and $s_t^k$ is the score of action $k$ for frame $t$.

To generate initial proposals from these actionness vectors, Algorithm 1 is used. Our basic assumption is that a snippet containing action should consist of frames with the actionness score higher than a threshold. Meanwhile, we observe that action snippets usually have limited durations, therefore the minimum and maximum frame lengths are used. Given the scores, a connected component scheme is devised to merge the neighboring regions with high scores (Line 14-32). Besides the original curve from the action scores, the smooth curve is also applied with Gaussian kernel density estimation (Line 8-9). After that, the proposal set is generated by the non-maximum suppression (NMS) of the candidate snippets.

\subsection{Refinement Network}

Refining the proposals is important to build an efficient action localization system. While the Actionness Network uses the frame-level information, the Refinement Network considers the context of the short video unit. A unit is represented as $U(s,d) = \{x_i|i=s,s+1,...,s+d\}$, where $X$ represents the video, $s$ is the starting frames and $d$ is unit length. The borders of the actions are usually vague. As a result, it is hard to train the model for the precise boundary regression. For each video, we generate units by cropping the proposal with a fixed length and stride. Any spatial or temporal features can be used to represent the unit. Here we employ the non-local pyramid features, which will be described in Section \ref{sec:ln}.

Context information will help the network to know how to start and where to end. To keep context information, we follow the pipeline in \cite{Zhao_2017_ICCV} and augment the range of unit $U$ from $[{s}, {s+d}]$ to $[{s-\frac{d}{2}}, {s+d+\frac{d}{2}}]$. As actions are usually composed by a set of motions, modeling the unit sequence is essential to the success of Refinement Network. To this end, we leverage an RNN-based sequence encoder. Specifically, we use Bi-directional Gated Recurrent Unit(BiGRU)~\cite{Schuster1997Bidirectional} as the RNN unit to encode the context information.

\begin{figure}[t]
  \centering
  \includegraphics[width=\linewidth]{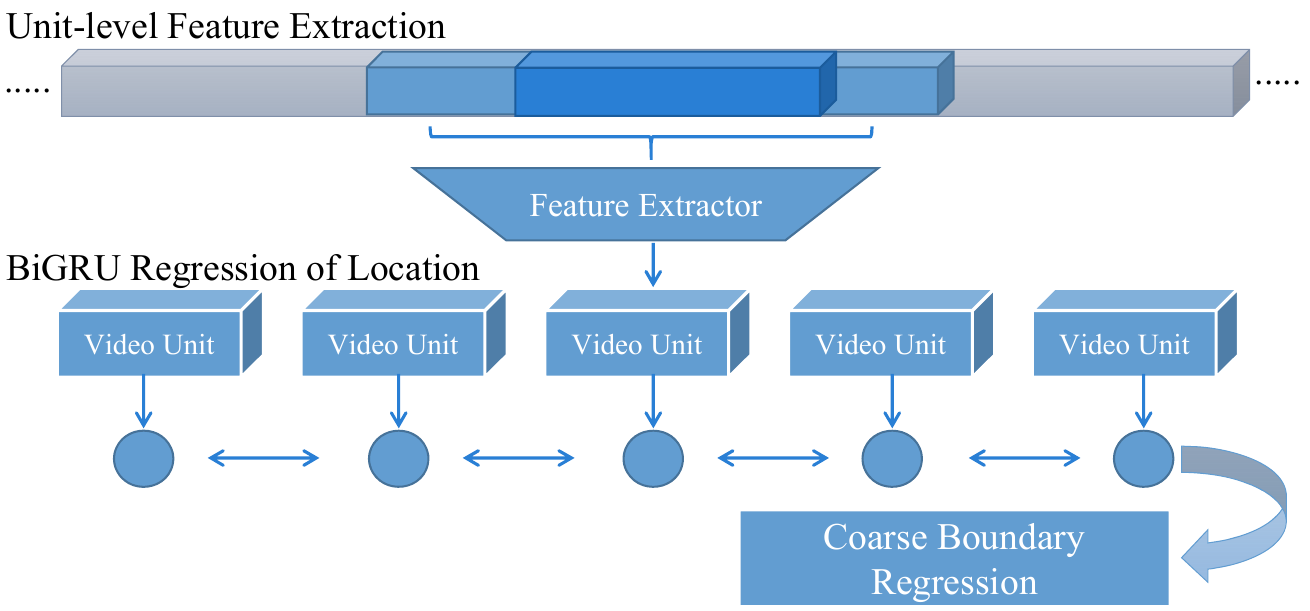}
  \caption{{Architecture of Refinement Network. The frame-level features are aggregated to get unit-level feature. Refinement Network walks through the unit-level features of the proposal to regress the coarse boundary.}}
  \label{fig:gru}
\end{figure}

GRU is the of architecture of recurrent unit that learns to encode a variable-length sequence into a fixed-length code. Figure \ref{fig:gru} illustrates the architecture of RN based on GRU. Rather than using five gates in LSTM, GRU has only two gates to control the updates of the hidden units, \ie, \textit{reset} gate $r$ and \textit{update} gate $z$.

The $i$-th hidden unit of \textit{reset} gate $r_j$ is computed by:
\begin{equation}
	r_j = \sigma([\mathrm{W}_rx]_j+[\mathrm{U}_rh^{\langle t-1\rangle}]_j)
\end{equation}
where $\sigma$ is sigmoid function and $[.]_j$ denotes the $j$-th element of a vector. The inputs are $x$ and previous state $h^{\langle t-1\rangle}$, and $\mathrm{W}_r$ and $\mathrm{U}_r$ are \textit{reset} gate weights.

The computation of \textit{update} gate $z$ is similar, which is:
\begin{equation}
	z_j = \sigma([\mathrm{W}_zx]_j + [\mathrm{U}_zh^{\langle t-1\rangle}]_j)
\end{equation}

The activation of hidden unit $h_j$ is computed by
\begin{equation}
	h_j^{\langle t\rangle} = z_jh_j^{\langle t-1\rangle}+(1-z_j)\hat{h}_j^{\langle t\rangle}
\end{equation}
where
$$
	\hat{h}_j^{\langle t\rangle} = \phi([\mathrm{W}x]_j+[\mathrm{U}(r\odot h^{\langle t-1\rangle}]_j)
$$

We define three types of proposals based on Intersection-over-Union (IoU) with its groundtruth: positive proposals with IoU $>$ 0.7, incomplete proposals with 0.3 $\leq$ IoU $<$ 0.7, and background proposals with IoU$<$0.1. During training, both input proposals and the target boundary coordinates are first fit to the applicable video unit. The unit-level features are encoded one by one through the BiGRU network. A full-connected layer then receives the output from BiGRU and is used to regress the localization.  {Parameter coordinate offset is used for regression. The Refinement Network regresses both interval center and proposal span. The regression loss is defined as

\begin{equation}\label{eq:loc}
    L_{loc} = \frac{1}{N}\sum_{i=1}^{N} \text{smooth}_{L_1}(c)+\text{smooth}_{L_1}(s),
\end{equation}

where $ {c} = \frac{c_{gt} - c_{unit}}{l_{unit}}$ is the interval center, ${s} = log\frac{l_{gt}}{l_{unit}}$ is the proposal span, $N=N_{pos}+N_{inc}$ }(positive and incomplete proposals are considered), and $\text{smooth}_{L_1}(\cdot)$ is the smooth-$L_1$ loss. Here $c_{y}$ and $l_{y}$ are the center frame and length of the groundtruth ($y=gt$) or video snippet ($y=unit$).

\subsection{Localization Network}
\label{sec:ln}

Localization Network is responsible for producing the evolved temporal proposals and the scores. Ideally, the network not only finds the recognized action and the precise action proposals, but also decides whether an proposal is the action or the background. To this end, we adopt the structured segment networks (SSN)~\cite{Zhao_2017_ICCV} as the backbone. Specifically, the temporal proposals from Refinement Network are augmented with three stages, \ie, \emph{starting}, \emph{course}, and \emph{ending}. The structured temporal pyramid features are then calculated in the SSN framework. Inspired by the application of non-local neural network for video classification~\cite{2017arxiv_xwang}, we extend the representation by adding a non-local block before the last layer. The non-local operation is defined as:
\begin{equation}\label{eq:nonlocal}
\mathbf{y}_i = \frac{1}{\mathcal{C(\mathbf{x})}} \sum_{\forall j}f(\mathbf{x}_i, \mathbf{x}_j)g(\mathbf{x}_j),
\end{equation}
where $\mathbf{x}$ and $\mathbf{y}$ are the input and the output signals respectively, $\forall j$ indicates the non-local behaviour for all positions in the frame, $g(\cdot)$ is the representation, and $f(\mathbf{x}_i, \mathbf{x}_j) = \theta(\mathbf{x}_i)^T\phi(\mathbf{x}_j)$ is the dot-product similarity function between $\mathbf{x}_i$ and $\mathbf{x}_j$. To avoid breaking the existing models, the non-local block is added with the residual connection~\cite{2016cvpr_khe}. We denote the representations as \emph{non-local pyramid features}. We will show in our experiments that this feature is superior to directly extracting the structured temporal pyramid features. Based on the features, the multi-task loss is used to train the Localization Network.

\vspace{0.12in}\noindent\textbf{Action Classification.} Positive and background proposals are used to train a classifier with $K$ classes for actions and 1 for background. We randomly sample the positive and background proposals to make the ratio around 1:1 to avoid data imbalance. Cross entropy loss is used as follows,
\begin{equation}
L_{cls} = \frac{1}{N_{tr}} \sum_{i=1}^{N_{tr}} - log(p_i^k)
\end{equation}
\begin{equation}\label{eq:pik}
p_i^k = \text{exp}(s_{i}^k)/\sum_j \text{exp}(s_{i}^j).
\end{equation}
where $N_{tr}$ is the number of training samples, and $s_{i}^j$ is the score of action $j$ for proposal $i$.

\vspace{0.12in}\noindent\textbf{Completeness Evaluation.} Only a few proposals will match the groundtruth instance. A binary classifier is used to predict a value $s_{comp}$ to represent whether the proposal is background or an action component, which plays an important role during evaluation in ranking the proposals. Positive and incomplete proposals are used to train this task. we use the online hard example mining strategy to overcome imbalance of dataset and improve classifier performance. We put positive proposals and incomplete proposals with a ratio of 1:4. During the training we only choose the first 1/4 incomplete proposal examples according to loss value to train loss with positive proposals. The Hinge loss is applied:
\begin{equation}
L_{comp} = \frac{1}{N_{tr}} \sum^{N_{tr}}_{i=1} \max(0,1-c_i \cdot p_i)
\end{equation}
where $c_i$ indicates whether the $i$-th proposal is positive proposal ($c_i = 1$) or incomplete proposal ($c_i = -1$), and $p_i$ is the probability that proposal $i$ is an action component.

\vspace{0.12in}\noindent\textbf{Localization Regression.} The loss function in Equation (\ref{eq:loc}) is used. Different from Refinement Network, this step only uses the positive proposals for boundary fine-tuning. Therefore, $N=N_{pos}$ is set for Equation (\ref{eq:loc}).

The overall multi-task loss function in Localization Network is defined as
\begin{equation}
L = L_{cls} + \alpha L_{comp} + \beta L_{loc}.
\end{equation}
We set $\alpha$ and $\beta$ to 0.1 in our experiments. Based on the multi-task loss, the normalized classification score $p^k_{cls}$ (see Equation (\ref{eq:pik})) and completeness score $s_{comp}$ are computed for a given proposal. The final ranking score for action $k$ is $s^k = p^k_{cls}\cdot\exp(s_{comp})$.

\begin{table}[t]
\centering
\caption{Action detection results on THUMOS14 (in \%). mAP at different IoU thresholds $\alpha$ are reported. The results are sorted in ascending order of the performances at $\alpha=0.5$. `-' indicates the results are not available in the corresponding papers. Bold faces are the top results while underlines correspond to the second runners-up.}
\begin{tabular}{l | c c c c c}
\hline
 IoU thresholds $\alpha$ & 0.3  & 0.4 & 0.5 & 0.6 & 0.7 \\ \hline
 Wang \etal~\cite{wang2014action} &  14.6 & 12.1 &  8.5 & 4.7  & 1.5 \\
 Oneata \etal~\cite{oneata2014lear} & 28.8  & 21.8 & 15.0 & 8.5  & 3.2 \\ \hline
 Heilbron \etal~\cite{caba2016fast} & -  & - & 13.5 & -  & - \\
 Escorcia \etal~\cite{escorcia2016daps} & -  & - & 13.9 & -  & - \\
 Richard and Gall~\cite{Richard_2016_CVPR} & 30.0  & 23.2 & 15.2& -  & -  \\ %\hline
 Yeung \etal~\cite{yeung2016end} & 36.0 & 26.4 & 17.1 & -  & - \\ %\hline
 PSDF~\cite{yuan2016temporal} &  33.6 & 26.1 & 18.8 & -  & - \\ %\hline
 S-CNN~\cite{shou2016temporal} & 36.3  & 28.7 & 19.0 & 10.3  & 5.3 \\
 Conv \& De-conv~\cite{shou2017cdc} & 38.6  & 28.2 & 22.4 & 12.0  & 7.5  \\
 CDC~\cite{shou2017cdc} & 40.1  & 29.4 & 23.3& 13.1  & 7.9  \\
 SSAD~\cite{2017mm_tlin} & 43.0 & 35.0 & 24.6 & - & - \\
 TPC+S-CNN~\cite{2017arxiv_kyang} & 41.9  & 32.5 & 25.3 & 14.7  & 9.0  \\
 TURN TAP~\cite{2017iccv_tap} & 44.1  & 34.9 & 25.6 & -  & - \\
 TPC+FGM~\cite{2017arxiv_kyang} & 44.1  & 37.1 & 28.2 & \underline{20.6}  & \underline{12.7}  \\
 TAG~\cite{2017arxiv_yxiong} & 48.7  & 39.8 & 28.2 & -  & -  \\
R-C3D~\cite{Xu_2017_ICCV} & 44.8  & 35.6 & 28.9 & -  & - \\
SSN~\cite{Zhao_2017_ICCV} & {\bf 51.9}  & 41.0 & 29.8 & -  & - \\
CBR~\cite{2017arxiv_jgao} & \underline{50.1}  & \underline{41.3} & \underline{31.0} & 19.1  & 9.9 \\
\hline
ETP~[ours] & 48.2  & {\bf 42.4} & {\bf 34.2} & {\bf 23.4}  & {\bf 13.9} \\ \hline
\end{tabular}
\label{tab:res_thumos14}
\end{table}

\section{Experiments}
\label{sec:exp}

In this section, we evaluate the effectiveness of the proposed framework on the action localization benchmarks. We first introduce the evaluation datasets and the experimental setup, and then evaluate the performance of the proposed framework and compare it with several state-of-the-art approaches. Finally, we discuss the effect of parameters and components. We denote our approach as {\bf ETP} ({\bf E}volving {\bf T}emporal {\bf P}roposals).

\subsection{Dataset and Evaluation Metric}

We conduct the experiments on the THUMOS Challenge 2014 dataset (THUMOS14) \cite{jiang2014thumos}. The whole dataset contains 1010 videos for validation and 1,574 videos for testing. For the temporal action localization task, only 200 validation-set videos and 213 testing-set videos with temporal annotations of 20 action classes are provided. The UCF101 dataset~\cite{soomro2012ucf101} is appointed as the official training set. As the videos in UCF101 are trimmed, we train our framework on the THUMOS14 validation set and then evaluate on the testing set.

The mean average precision (mAP) at different IoU thresholds $\alpha$ are reported. Following recent works for precise temporal action localization~\cite{shou2017cdc,2017arxiv_kyang}, we consider the thresholds $\alpha$ of \{0.3, 0.4, 0.5, 0.6, 0.7\}, and the mAP at $\alpha=0.5$ is used for comparing different approaches.

\begin{figure*}[t]
  \centering
  \includegraphics[width=.95\linewidth]{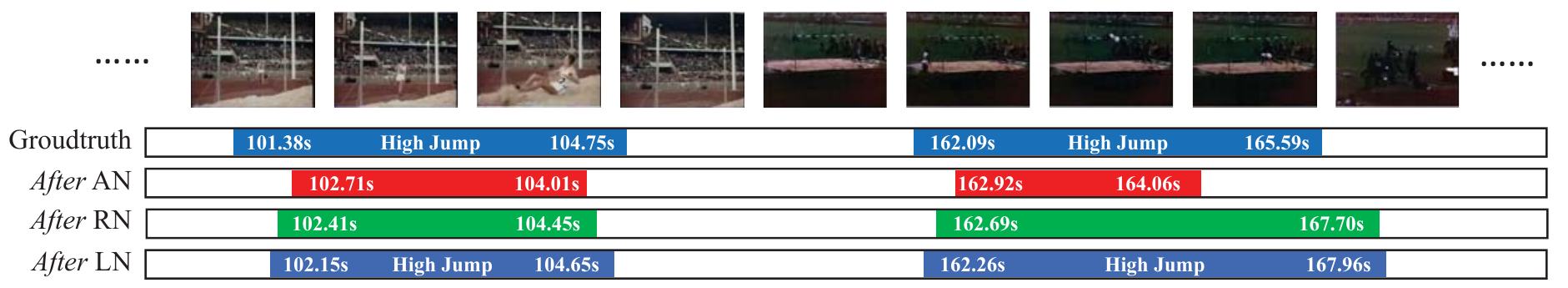}\\
  \vspace{0.13in}
  \includegraphics[width=.95\linewidth]{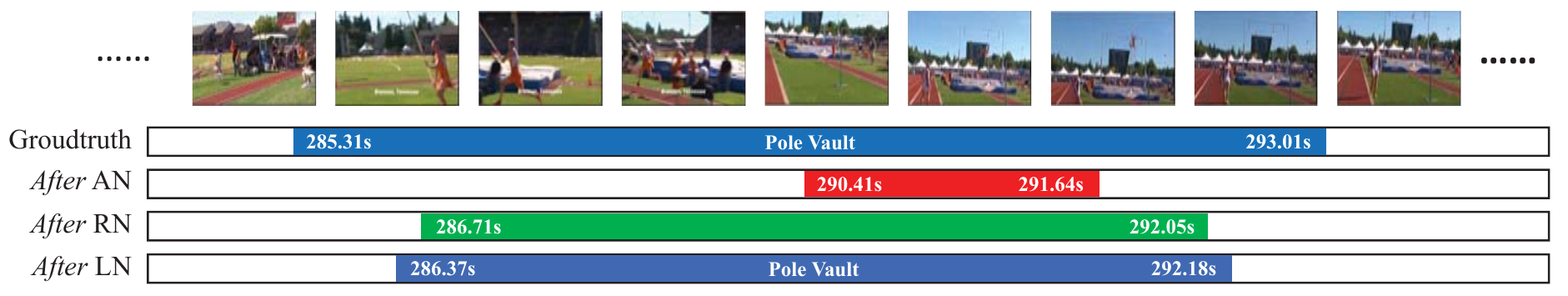}\\
  \vspace{0.13in}
  \includegraphics[width=.95\linewidth]{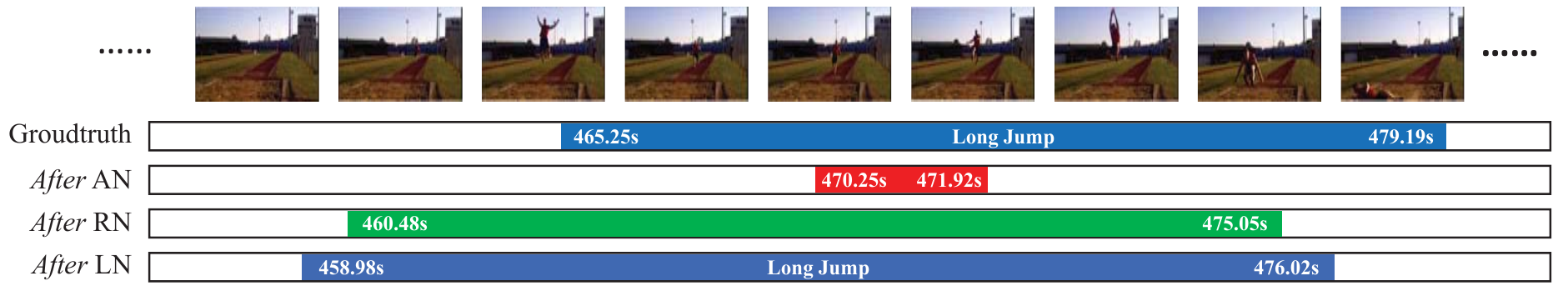}
  \caption{Visualization of the action instances by our proposed approach on THUMOS14 dataset.}
  \label{fig:thumos_res}
\end{figure*}

\subsection{Experimental Settings}

\noindent\textbf{Training.} We implment our approach based on PyTorch\footnote{\url{http://pytorch.org/}}. In the Actionness Network, we use the ResNet-34~\cite{2016cvpr_khe} to extract the frame-level class-specific actionness scores. The ResNet-34 network is pre-trained on ImageNet dataset and then fine-tuned on the UCF101 dataset. Random horizontal flip and random center crop are used for data augmentation.

We use 2 BiGRU cells with 512 hidden values as RNN units to build the Refinement Network. The training batch size is 128 and it has 20$K$ iterations. Only the positive and incomplete proposals are used as the training samples. We use SGD as the optimizer to train the network with momentum 0.9. The base learning rate of the network is $0.1$ with the decay rate $0.1$ to decrease the learning rate at every 5$K$ iterations.

In Localization Network, the feature extraction is based on Inception V3~\cite{InceptionV3} with batch normalization pre-trained on Kinetics Human Action Video dataset~\cite{kineticsdataset}. Both spatial and temporal flow networks are trained using SGD with momentum 0.9. The network is trained for 90$K$ iterations with the learning rate of 0.1 and scaled down by 0.1 every 5$K$ iterations until the learning rate is less than $10^{-5}$.

\vspace{0.12in}
\noindent\textbf{Inference.} During testing phase, the initial proposals are generated by Algorithm 1. The Refinement Network is then used for coarse regression of the proposals, which are sent to Localization Network to enhance the boundary regression and predicted action classes. The choice of NMS threshold has important influences on testing results. To achieve precise localization results with higher IoU thresholds, we empirically set the NMS threshold to 0.36.

\begin{table}[t]
\centering
\caption{Per-class AP at $\alpha=0.5$ on THUMOS14 (in \%).}
 \begin{tabular}{l| c c c c}
 \hline
 ~ & \cite{yeung2016end} &   \cite{shou2016temporal} & \cite{Xu_2017_ICCV} & Ours\\ \hline
 BaseballPitch  & 14.6  & 14.9&  \bf{26.1} & \underline{22.5} \\ %\hline
 BasketballDunk & 6.3   &20.1 & \bf{54.0} & \underline{30.3} \\ %\hline
 Billiards & \bf{9.4} &  7.6 & \underline{8.3} & 8.1 \\ %\hline
 CleanAndJerk  & \bf{42.8}   &24.8 & 27.9 & \underline{40.9} \\ %\hline
 CliffDiving    & 15.6   &\underline{27.5} & \bf{49.2} & 16.7 \\ %\hline
 CricketBowling & 10.8   & 15.7& \bf{30.6} & \underline{16.3} \\ %\hline
 CricketShot    & 3.5   &\bf{13.8} & \underline{10.9} & 7.2 \\ %\hline
 Diving & 10.8    &   17.6 &\underline{26.2} & \bf{50.9} \\ %\hline
 FrisbeeCatch   & 10.4   &\underline{15.3} & \bf{20.1} & 2.3 \\ %\hline
 GolfSwing      &13.8    &\underline{18.2} & 16.1 & \bf{44.4} \\ %\hline
 HammerThrow    & 28.9 &  19.1 &\underline{43.2} & \bf{71.7} \\ %\hline
 HighJump       & \underline{33.3} &  20.0 & 30.9 & \bf{51.2}\\ %\hline
 JavelinThrow   & 20.4 &  18.2 & \underline{47.0} & \bf{47.3} \\ %\hline
 LongJump       & 39.0 &  34.8 & \underline{57.4} & \bf{81.9} \\ %\hline
 PoleVault      & 16.3 &  32.1 & \underline{42.7} & \bf{56.5} \\ %\hline
 Shotput & 16.6  &  12.1 & \underline{19.4} & \bf{32.0} \\ %\hline
 SoccerPenalty  & 8.3  &\underline{19.2} & 15.8 & \bf{19.7} \\ %\hline
 TennisSwing    & 5.6  &\underline{19.3} & 16.6 & \bf{29.1} \\ %\hline
 ThrowDiscus    &29.5  &24.4 & \underline{29.2} & \bf{39.1}\\ %\hline
 VolleyballSpiking & 5.2 &  4.6 &\underline{5.6} & \bf{14.0}\\ \hline
 mAP@0.5 &17.1 & 19.0 & \underline{28.9} & \bf{34.2}\\ \hline
 \end{tabular}
\label{tab:per_class_ap}
\end{table}

\subsection{Comparison with State-of-the-arts}

We first evaluate the overall results of our proposed framework for action localization and compare them with several state-of-the-art approaches. There are a few parameters in ETP, including the lengths of units in the Refinement Network, and feature types in the Actionness Network and the Localization Network. In our experiments, we set 64 frames for unit length, and extract the non-local pyramid features. The effects of these parameters, as well as the components of the framework, will be evaluated in Section \ref{sec:as}.

Table \ref{tab:res_thumos14} summarizes the mAPs of all action classes in THUMOS14. We compare ETP with the results during the challenge~\cite{wang2014action,oneata2014lear} and state-of-the-art approaches. From Table \ref{tab:res_thumos14} we can see that when $\alpha=0.5$, ETP outperforms all the challenge results as well as the state-of-the-art approaches shown in the middle part of Table \ref{tab:res_thumos14}, including the actionness based approach TAG~\cite{2017arxiv_yxiong}, the Convolutional-De-Convolutional Networks~\cite{shou2017cdc}, the Cascaded Boundary Regression models~\cite{2017arxiv_jgao}, and a very recent Temporal Preservation Networks~\cite{2017arxiv_kyang}. The substantial performance gains over the previous works under different IoU thresholds confirm the effectiveness of our evolving temporal proposals for precise temporal action localization. Table \ref{tab:per_class_ap} further shows the per-class results at $\alpha=0.5$ for our approach and several previous works (Yeung \etal~\cite{yeung2016end}, S-CNN~\cite{shou2016temporal}, and R-C3D~\cite{Xu_2017_ICCV}). Notably, our approach performs the best on 12 action classes and shows significant improvement (by more than 20\% absolute AP over the next best) for 10 actions such as Diving, High Jump, Pole Vault, and Long Jump. Figure \ref{fig:thumos_res} illustrates some prediction results for these actions respectively.

\begin{figure*}[t]
  \centering
  \includegraphics[width=.82\linewidth]{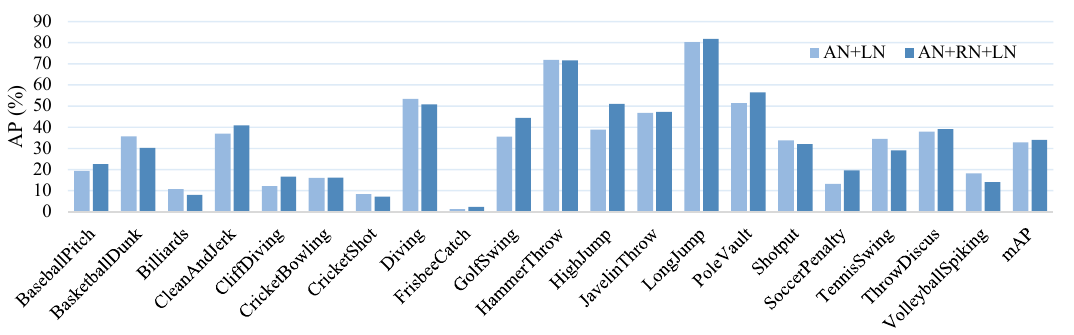}
  \caption{Per-class AP@$\alpha=0.5$ with incorporating of Refinement Network.}
  \label{fig:component}
\end{figure*}
\begin{figure*}[t]
  \centering
  \includegraphics[width=.82\linewidth]{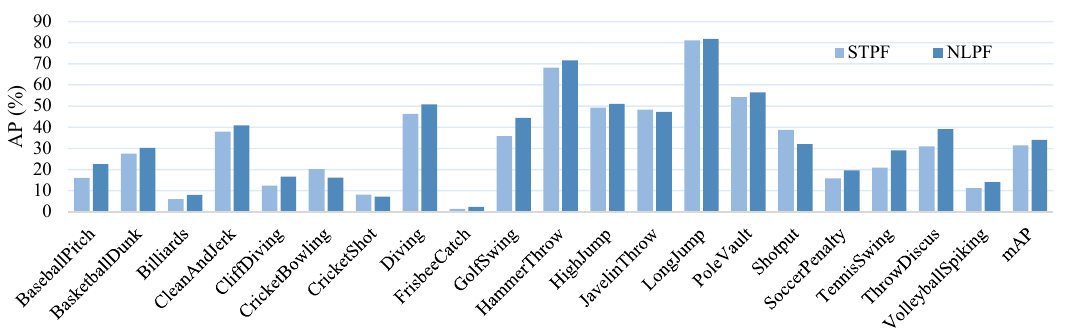}
  \caption{Per-class AP@$\alpha=0.5$ w.r.t. various features (STPF: structured temporal pyramid features; NLPF: non-local pyramid features).}
  \label{fig:features}
\end{figure*}
\begin{figure}[t]
  \centering
  \includegraphics[width=.85\linewidth]{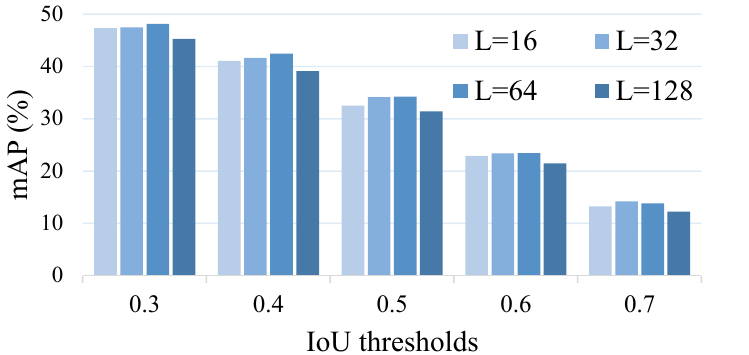}
  \caption{Evaluation of the ETP framework with different video unit lengths in Refinement Network on THUMOS14.}
  \label{fig:cliplen}
\end{figure}

\subsection{Ablation Study}
\label{sec:as}

In this part, the control experiments of switching off different components of the proposed framework are conducted.

\vspace{0.12in}
\noindent\textbf{Effect of Refinement Network} We first evaluate the effect of the Refinement Network over the whole framework. Figure \ref{fig:component} gives the comparison of all the action classes with the pipeline that the outputs of Actionness Network are directly sent to Localization Network. There are 13 categories with performance gains after incorporating with Refinement Network and the overall AP is increased by 1.3\%.

\vspace{0.12in}
\noindent\textbf{Unit Length in RN.} We also evaluate the setting of the video unit length in the Refinement Network. The lengths of \{16,32,64,128\} are tested, and the stride is set to half of the length. In Figure \ref{fig:cliplen} we plot the localization results versus the video unit length and the IoU thresholds $\alpha$. Using the length of 16 frames per video unit, we can already get the mAP of 32.6\% when IoU threshold $\alpha=0.5$, which already outperforms previous works. We observe significant performance gains when the length increases from 16 to 32, after which the performance tends to be saturated. There is a performance drop when the unit length is 128, and we conjecture that a larger unit length may not be the one at which the GRU-based sequence encoder responds with optimal context information.

\vspace{0.12in}
\noindent\textbf{Non-local Pyramid Features.} Recall that the non-local pyramid features by adding a non-local residual block, based on the non-local operation shown in Equation (\ref{eq:nonlocal}). In Figure \ref{fig:features} we report the detailed localization results for each action learned using the non-local pyramid features as well as the structured temporal pyramid features in \cite{Zhao_2017_ICCV}. For 16 action classes we achieve higher AP values, which is consistent with the observations from previous work~\cite{2017arxiv_xwang}.

\begin{table}[t]
\centering
\caption{mAP from different modalities.}
\begin{tabular}{l | c c c c c}
\hline
  IoU thresholds $\alpha$ & 0.3  & 0.4 & 0.5 & 0.6 & 0.7 \\ \hline
  RGB & \underline{39.9}  & \underline{33.7} & \underline{25.3} & 16.1 & 8.6\\
  Flow & 34.8  & 30.4 & 25.0 & \underline{18.0} & \underline{10.7} \\
  \hline
  RGB+Flow & \bf{48.2}  & \bf{42.4} & \bf{34.2} & \bf{23.4}  & \bf{13.9} \\
  \hline
\end{tabular}
\label{tab:rgb_flow}
\end{table}

\vspace{0.12in}
\noindent\textbf{Video Modality.} Our last experiment evaluates the effect of different modalities for temporal action localization. The results are shown in Table \ref{tab:rgb_flow}. The RGB modality achieves a higher mAP when the IoU threshold $\alpha$ is smaller, and the Flow modality tends to be better after increasing $\alpha$. As shown in Table \ref{tab:rgb_flow}, using both modalities leads to performance gains for all IoU thresholds.

\section{Conclusions}
\label{sec:conclusion}

In this paper, we have proposed the Evolving Temporal Proposals (ETP), a framework with three components (\ie, Actionness Network, Refinement Network, and Localization Network) to generate temporal proposals for precise action localization in the untrimmed videos. Through empirical temporal action localization experiments, we have shown that ETP is more effective than previous systems by generating very competitive results on the THUMOS14 dataset. We leverage the non-local pyramid features to effectively model the activity, which improves the discriminative ablity between completeness and incompleteness of a proposal. For future work, we plan to optimize the inference procedure of the proposed framework, and explore the one-stream video action localization.

\bibliographystyle{ACM-Reference-Format}
\balance
\bibliography{arxiv}
\end{document}